\documentclass[accepted]{uai2021} 

\usepackage[american]{babel}

\usepackage{natbib} 
\usepackage{mathtools} 
\usepackage{booktabs} 
\usepackage{tikz} 

\usepackage[ruled,vlined,linesnumbered,noend]{algorithm2e}
\usepackage{amsmath}
\usepackage{amsfonts}
\usepackage{enumitem}
\usepackage{subfigure}
\usepackage{soul}
\newcommand{\squishlist}[1][$\bullet$]
{
    \begin{list}{#1}
    {
        \setlength{\itemsep}{0pt}
        \setlength{\parsep}{2pt}
        \setlength{\topsep}{2pt}
        \setlength{\partopsep}{0pt}
        \setlength{\leftmargin}{1.5em}
        \setlength{\labelwidth}{1.5em}
        \setlength{\labelsep}{0.5em}
    }
}

\newcommand{\squishend}{\end{list}}

\title{CLAIM: Curriculum Learning Policy for Influence Maximization in Unknown Social Networks}

\author[1]{\href{mailto:<dexunli.2019@phdcs.smu.edu.sg>?Subject=Your UAI 2021 paper}{Dexun Li}{}}
\author[1]{Meghna Lowalekar}
\author[1]{Pradeep Varakantham}
\affil[1]{%
    School of Computing and Information Systems\\
    Singapore Management University\\
    Singapore
}
  
\begin{document}
\maketitle

\begin{abstract}
    Influence maximization is the problem of finding a small subset of nodes in a network that can maximize the diffusion of information. Recently, it has also found application in HIV prevention, substance abuse prevention, micro-finance adoption, etc., where the goal is to identify the set of peer leaders in a real-world physical social network who can disseminate information to a large group of people. Unlike online social networks, real-world networks are not completely known, and collecting information about the network is costly as it involves surveying multiple people. In this paper, we focus on this problem of network discovery for influence maximization. The existing work in this direction proposes a reinforcement learning framework. As the environment interactions in real-world settings are costly, so it is important for the reinforcement learning algorithms to have minimum possible environment interactions, i.e, to be sample efficient. In this work, we propose CLAIM - \textbf{C}urriculum \textbf{L}e\textbf{A}rning \textbf{P}olicy for \textbf{I}nfluence \textbf{M}aximization to improve the sample efficiency of RL methods. We conduct experiments on real-world datasets and show that our approach can outperform the current best approach.
\end{abstract}

\section{Introduction}\label{sec:intro}
Social interactions between people play an important role in spreading information and behavioral changes. The problem of identifying a small set of influential nodes in a social network that can help in spreading information to a large group is termed as influence maximization (IM) \citep{Kempe03}. It was widely used in applications such as viral marketing \citep{Kempe03}, rumor control \citep{Budak11}, etc, which use the online social network. In addition to these, IM has also found useful applications in domains involving real world physical social networks. Some of these applications include identifying peer leaders in homeless youth network to spread awareness about HIV \citep{Wilder18,yadav2016using}, identifying student leaders in school network to disseminate information on substance abuse \citep{valente2007identifying}, identifying users who can increase participation in micro-finance \citep{banerjee2013diffusion}, etc. In the case of real world social networks, the network information is not readily available and it is generally gathered by individually surveying different people who are part of the network. As conducting such surveys is a time intensive process requiring  substantial efforts from a dedicated team of social work researchers, it is not practically possible to have access to a complete network structure. Therefore, the influence maximization problem in the real world is coupled with the uncertain problem of discovering network using a limited survey budget (i.e., the number of people who can be queried).

Most of the existing work \citep{Wilder18,Bryan18,yadav2016using} which addresses real-world influence maximization problems perform network discovery by surveying nodes while exploiting a specific network property such as community structure. CHANGE algorithm \citep{Wilder18} is based on the principle of \textit{friendship paradox} and performs network discovery by surveying a random node and one of its neighbour. Each node reveals the information about its neighbors upon querying. The subgraph obtained after querying a limited set of nodes is used to pick a set of influential nodes using an influence maximization algorithm. A recent work by \citet{kamarthi2019influence} provides a reinforcement learning based approach to automatically train an agent for network discovery. They developed an extension to DQN referred to as Geometric-DQN to learn policies for network discovery by extracting relevant graph properties, which achieves better performance than the existing approaches. As any other reinforcement learning approach, the work by \citet{kamarthi2019influence} needs to perform multiple interactions with the environment to perform exploration. As in the real world settings, the environment interactions are costly, the approach can be improved by reducing the environment interactions, i.e., by increasing the sample efficiency. This approach employs a myopic heuristic (new nodes discovered) to guide exploration and we employ goal directed learning to provide a forward looking (non-myopic) heuristic. 

In this work, we propose to model the network discovery problem as a goal-directed reinforcement learning problem. We take the advantage of the \textit{Hindsight Experience Replay} \citep{Andrychowicz2017} framework which suggests learning from failed trajectories of agent by replaying each episode with a different goal (e.g. the state visited by agent at the end of its failed trajectory) than the one agent was trying to achieve. This helps in increasing sample efficiency as agent can get multiple experiences for learning in a single environment interaction. To further improve the performance, we use the curriculum guided selection scheme proposed by \citet{Fang2019} to select the set of episodes for experience replay. While there have been some other works which focus on improving the sample-efficiency \citep{sukhbaatar2017intrinsic,burda2018large,colas2019curious}, most of them are designed for domain specific applications and unlike our curriculum guided selection scheme which adaptively controls the exploration-exploitation trade-off by gradually changing the preference on goal-proximity and diversity-based curiosity, they only perform curiosity-driven learning.

\noindent \textbf{Contributions:} In summary, following are the main contributions of the paper along different dimensions:

\begin{itemize}[leftmargin=*]
\item  \textbf{Problem:} We convert the whole process of network discovery and influence maximization into a goal directed learning problem. Unlike standard goal directed learning problems where goal state is known, in this problem, the goal state is not given. We provide a novel heuristic to generate goals for our problem setting.

\item \textbf{Algorithm:} We propose a new approach CLAIM - \textbf{C}urriculum \textbf{L}e\textbf{A}rning Policy for \textbf{I}nfluence \textbf{M}aximization in unknown social networks which by using Curriculum guided hindsight experience replay and goal directed Geometric-DQN architecture can learn sample efficient policies for discovering network structure. 

\item 
\textbf{Experiments:} We perform experiments in social networks from three different domains and show that by using our approach, the total number of influenced nodes can be improved by upto 7.51\% over existing approach.

\end{itemize}
\begin{table}[htp]
{\small
\begin{center}
\begin{tabular}{|l|l|}
\toprule
\textbf{Notation} & \textbf{Description}\\
\hline
$G^{\ast}=(V^{\ast},E^{\ast})$ & Entire Unknown Graph\\
\hline
$S$ & Set of nodes known initially \\
\hline 
$G_{t} = (V_{t},E_{t})$ &  Subgraph of $G^{\ast}$ discovered after $t$ queries\\
\hline
$N_{G^{\ast}}(u)$ &  Neighbors of vertex $u$ in graph $G^{\ast}$  \\
\hline
$E(X,Y)$ &All direct edges that connect a node in \\
~& set $X$ and a node in set $Y$\\
\hline
$O(G)$ & Set of nodes from graph $G$ selected by \\
~& influence maximization algorithm $O$\\
\hline
$I_{G^{\ast}}(A)$ & Expected Number of nodes influenced in \\
~&                   graph$G^{*}$ on choosing $A$ as the set of \\
~&                     nodes to activate\\
\bottomrule
\end{tabular}
\end{center}
}
\vspace{-0.1in}
\caption{ Notations}
\label{table:notations}
    \vskip -10pt
\end{table}

\section{Problem Description}\label{sec:prob}
The problem considered in this work involves discovering a subgraph of the unknown network such that the set of peer leaders chosen from the discovered subgraph maximizes the number of people influenced by peer leaders. We now describe both the components of the problem, i.e., network discovery and influence maximization in detail. The notations used in the problem description are defined in Table \ref{table:notations}.

\begin{itemize}
    \item \textbf{Network Discovery Problem:} The network discovery problem can be described as a sequential decision making problem where at each step, the agent queries a node from the discovered subgraph. The queried node reveals its neighbors, expanding the discovered subgraph. The process goes on for a fixed number of steps, determined by the budget constraint. Formally, initially we are given a set of nodes $S$ and the agent can observe all the neighbors of nodes in set $S$. Therefore, $V_{0} = S \cup N_{G^{\ast}}(S)$. The agent has a budget of $T$ queries to gather additional information. For $(t+ 1)^{th}$ query, the agent can choose a node $u_t$ from $G_t$ and observe $G_{t+1}$.
\begin{equation*}
\resizebox{.99\linewidth}{!}{
      $G_{t+1} = (V_t\cup N_{G^\ast}(u_t),E(G_t) \cup E(N_{G^\ast}(u_t),\{u_t \}))$.
      }
\end{equation*}

\end{itemize}
At the end of network discovery process, i.e., after $T$ queries, we get the final discovered subgraph $G_T$. This graph is provided as an input to an IM algorithm.
\begin{itemize}
    \item \textbf{Influence Maximization (IM) :}  IM is the problem of choosing a set of influential nodes in a social network who can propagate information to maximum nodes. In this paper, the information propagation over the network is modeled using Independent Cascade Model (ICM ) \citep{Kempe03}, which is the most commonly used model in the literature. In the ICM, at the start of the process, only the nodes in the set of chosen initial nodes are active. The process unfolds over a series of discrete time steps, where at every step, each newly activated node attempts to activate each of its inactive neighbors and succeeds with some probability $p$. The process ends when there are no newly activated nodes at the final step.  
    After discovering the subgraph $G_T$ using network discovery process, we can use any standard influence maximization algorithm to find out the best set of nodes to activate based on the available information. \citet{lowalekar2016robust} showed the robustness of the well-known greedy approach \citep{Kempe03} on medium scale social network instances, which is also served as the oracle in our paper.
\end{itemize}

\begin{figure}
  \centering
  \includegraphics[width=\linewidth,height=1.8in]{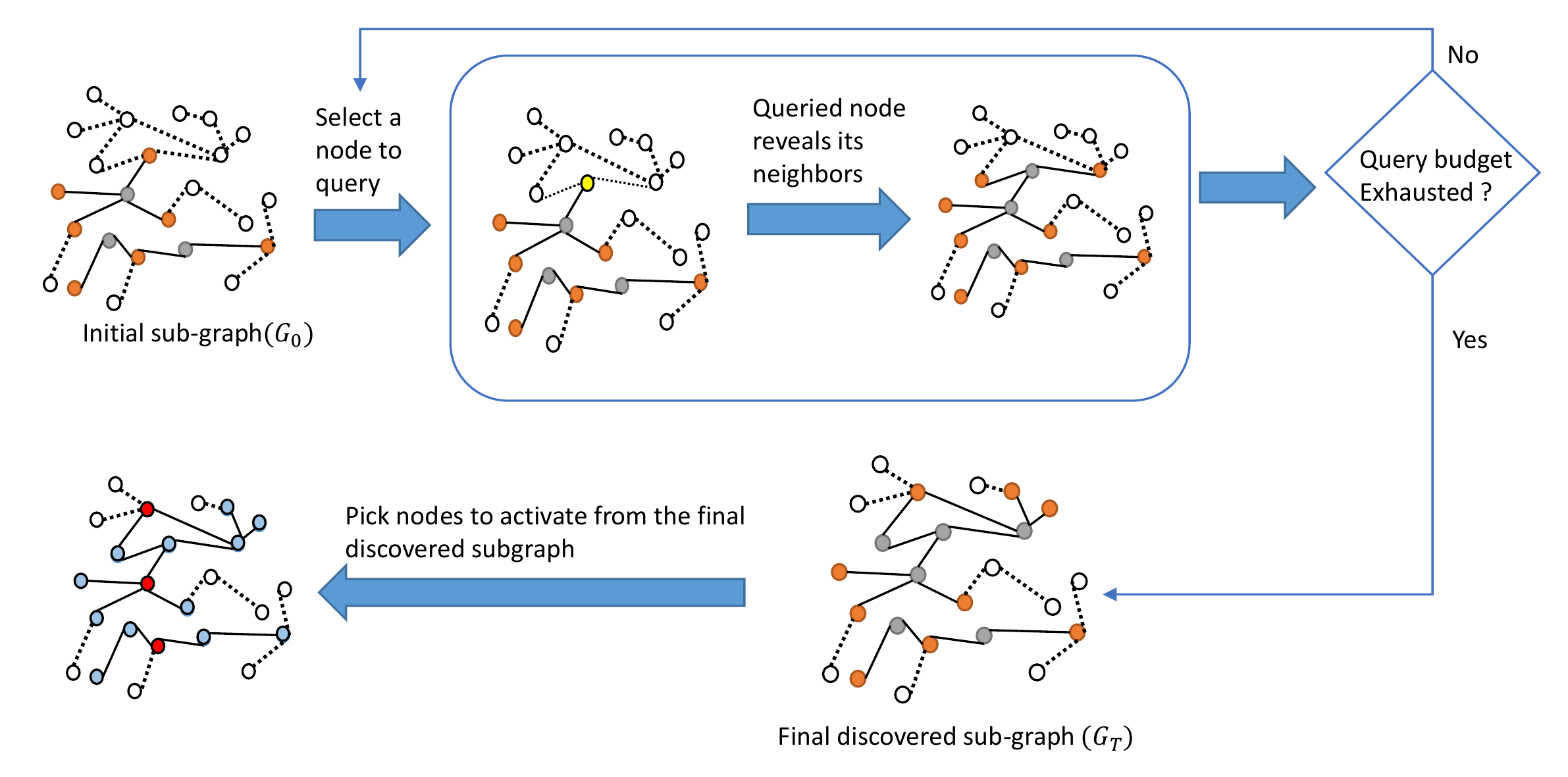}
  \caption{{\small Network discovery and influence maximization. Grey: Set of Queried Nodes; Orange: Set of unqueried nodes (In the initial subgraph $G_{0}$, grey nodes will represent the set $S$ and orange nodes will represent the set $N_{G^{\ast}}(S)$), Yellow: node picked by agent to query ($u_{t}$); Red: nodes selected by influence maximization algorithm in the final discovered subgraph  ($O(G_{T})$); Blue: other nodes in the final discovered subgraph $G_{T}$} }\label{fig:problem}
\end{figure}
Overall, given a set of initial nodes $S$ and its observed connections $N_{G^\ast}(S)$, our task is to find sequence of queries $(u_0,u_1,\dots,u_{T-1})$ such that $G_T$ maximizes $I_{G^\ast}(O(G_T))$.  Figure \ref{fig:problem} shows the visual representation of the problem.

\section{Background}\label{sec:bg}
In this section, we describe the relevant research, the MDP formulation and the Geometric-DQN architecture used by \citet{kamarthi2019influence} to solve the network discovery and influence maximization problem. 

\subsection{MDP Formulation}
\label{sect:mdp}
The social network discovery and influence maximization problem can be formally modelled as an MDP.
\begin{itemize}
\item {\bf State:} The current discovered graph $G_t$ is the state.
\item {\bf Actions:} The nodes yet to be queried in network $G_t$ constitute the action space. So, set of possible actions is $V_t\setminus \{ S\cup_{i\leq t} u_i\} ~ \forall t>0 \text{ and } N_{G^\ast}(S) \text{ when } t = 0$.
\item {\bf Rewards:}  Reward is only obtained at the end of episode, i.e,, after T steps. It is the number of nodes influenced in the entire graph $G^{*}$ using $G_{T}$, i.e., $I_{G^{*}}(O(G_{T}))$. The episode reward is denoted by $R_{T}$, where $T$ is the length of the episode (budget on the number of queries available to discover the network). 
\end{itemize}

\noindent \textbf{Training:} To train the agent in the MDP environment, DQN algorithm is used but the original DQN architecture which takes only the state representation as an input and outputs the action values can not be used as the action set is not constant and depends on the current graph. Therefore, both state and action are provided as an input to DQN and it predicts the state action value. The DQN model can be trained using a single or multiple graphs. If we train simultaneously on multiple graphs, then the MDP problem turn out to be Partially observable MDP, as the next state is determined by both current state and current action as well as the graph we are using. The range of reward values also depends on the size and structure of the graph, therefore, the reward value is normalized when multiple graphs are used for training. 
\begin{equation}
    R_{T} = \frac{I_{G^{*}}(O(G_{T}))}{OPT(G^{*})}
\end{equation}

\subsection{Geometric-DQN}
\begin{figure}
  \centering
  \includegraphics[width=\linewidth,height=1.8in]{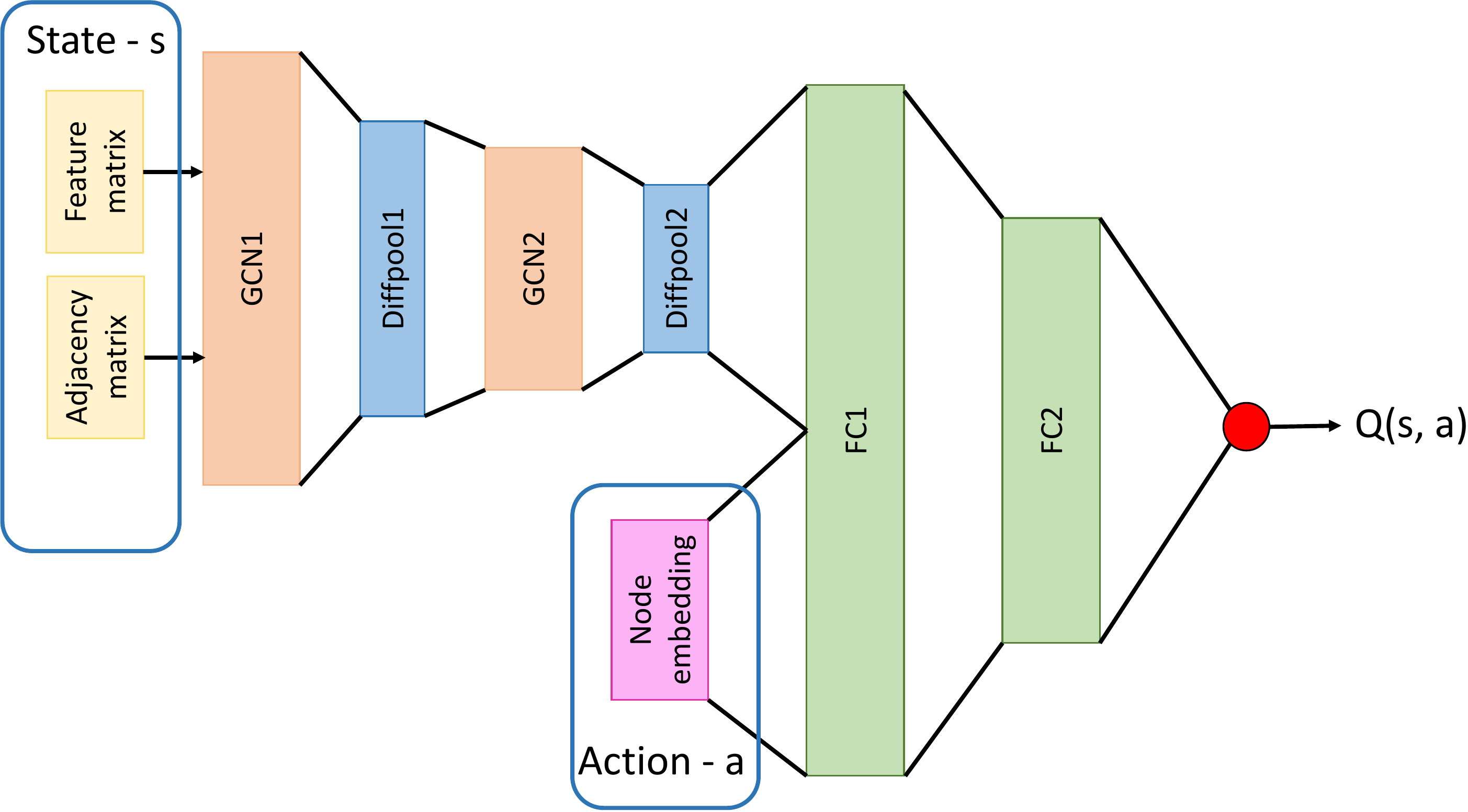}
  \vspace{-0.1in}
  \caption{{\small Geometric-DQN Architecture. Figure taken from \citet{kamarthi2019influence}. FC1/FC2 - fully connected layers.}} \label{fig:gcn-arch}
  \vspace{-0.1in}
\end{figure}

As described in previous section, the state is the current discovered graph $G_{t}$ and actions are the unqueried nodes in the current discovered graph. So, a good vector representation of the current discovered graph is required. It is also important to represent nodes such that it encodes the structural information of the node in the context of the current discovered graph. Figure~\ref{fig:gcn-arch} shows the Geometric-DQN architecture which takes the state and action representation as input and outputs the $Q(s,a)$ values. The details about state and action representation are provided below. 

\begin{itemize}[leftmargin=*]
\item  \textbf{State representation:} The state is the current graph. and the Geometric-DQN architecture uses Graph Convolutional Networks to generate graph embeddings. The graph $G_{t}$ is represented with the adjacency matrix $A_t \in \mathbb{R}^{|V_{t}|\times |V_{t}|}$ and a node feature matrix  $F_{t}^{(k-1)}\in\mathbb{R}^{|V_{t}| \times d}$ in layer $k-1$ where $d$ is the number of features. The node features in the input layer of graph convolution network, i.e., $F_{t}^{0}$ are generated by using random-walk based Deepwalk embeddings\footnote{Deepwalk learns node representations that are similar to other nodes that lie within a fixed proximity on multiple random walks.} \citep{Perozzi14}.

Now, a Graph Convolutional layer derives node features using a transformation function $F^{k} = M(A,F^{k-1};W^{k})$, where $W^{k}$ represent the weights of the $k^{th}$ layer. Using the formulation in \citet{Ying19}, the transformation function is given by
\begin{align}
    F_{t}^{(k)} &= ReLU(D^{-\frac{1}{2}} \tilde{A} D^{-\frac{1}{2}} F_{t}^{(k-1)}W^{(k)}) \nonumber
\end{align}
where $\tilde{A}$ means adjacency matrix $A_{t}$ with added self-connections, i.e., $\tilde{A} = A_{t} + I_n$  ($I_n$ is the identity matrix). $D=\sum_j \tilde{A}_{ij}$. To better represent the global representation of graph, differential pooling is used which learns hierarchical representations of the graph in an end-to-end differentiable manner by iteratively coarsening the graph, using graph convolutional layer as a building block. The output of graph convolutional network is provided as an input to a pooling layer.

\item \textbf{Actions representation:} DeepWalk node embeddings are also utilized for representing actions. We use $\phi$ to denote the deepwalk embeddings. 
\end{itemize}

Therefore, if $G_{t}$ is the current graph (state) and $u_{t}$ is the current node to be queried (action), we represent state as $S_{t} = (F_{t}^{0},A_{t})$ and action as $\phi(u_{t})$ which are input to the network as shown in the Figure~\ref{fig:gcn-arch}.

\section{Our Approach - CLAIM}
In this section, we present our approach CLAIM - \textbf{C}urriculum \textbf{L}e\textbf{A}rning Policy for \textbf{I}nfluence \textbf{M}aximization in unknown social networks. We first explain how the problem can be translated into a Goal directed learning problem. The advantage of translating the problem into goal directed learning problem is that it allows us to increase sample efficiency by using the Curriculum guided Hindsight experience replay (CHER) \citep{Fang2019}. CHER involves replaying each episode with  pseudo goals, so the agent can get multiple experiences in a single environment interaction which results in increasing the sample efficiency.

To use goal directed learning in our setting, we first present our novel heuristic to generate goals and the modifications to the MDP formulation for goal directed learning. After that, we present our algorithm to generate curriculum learning policy using Hindsight experience replay.

\subsection{Goal Directed Reinforcement Learning}
In the Goal Directed or Goal Conditioned Reinforcement Learning \citep{Andrychowicz2017,nair2018visual}, an agent interacts within an environment to learn an optimal policy for reaching a certain goal state or a goal defined by a function on the state space in an initially unknown or only partially known state space. If the agent reaches the goal, then the reinforcement learning method is terminated, and it solves the goal-directed exploration problem. 

In these settings, the reward that agent gets from the environment is also dependent on the goal that agent is trying to achieve. A goal-conditioned $Q$-function $Q(s, a, g)$ \citep{schaul15} learns the expected return for the goal $g$ starting from state $s$ and taking action $a$. Given a state $s$, action $a$, next state $s'$ , goal $g$ and corresponding reward $r$, one can train an approximate $Q$-function parameterized by $\theta$ by minimizing the following Bellman error:
\begin{align}
\frac{1}{2} || Q_\theta(s,a,g) - (r+\gamma \cdot \max_{a'} Q_{\theta'}(s',a',g)||^2 \nonumber
\end{align}
This loss can be optimized using any standard off-policy reinforcement learning method \citep{nair2018visual}.

Generally, in these goal-directed reinforcement learning problems, a set of goal states or goals defined by a function on the state space is given and the agent needs to reach one of the goal states (goals). But in our setting, we do not have an explicit goal state given. To convert the network discovery and influence maximization problem to a goal directed learning problem, we introduce the notion of goals for our problem. We define goal as the expected long term reward, i.e., the expected value of the number of nodes which can be influenced in the network and any state which can achieve this goal value becomes our goal state. As we have limited query budget to discover the network, the goal value will be highly dependent on the initial sub-graph. If we use the same value of goal for each start state, for some start states this common goal value will turn out to be a very loose upper bound (or very loose lower bound). Experimentally, we found that if the goal value is too far from the actual value which can be achieved, it negatively affects the speed of learning. So, we design a heuristic to compute a different goal for each start state.

\subsubsection{Goal Generation Heuristic}
As we need to generate goal at the start of each episode (i.e., before the agent starts interacting with the environment), we need to compute the goal value without making any queries to the environment. We assume that based on the domain knowledge, agent can get an estimate about the number of nodes ($|\tilde{V}^{\ast}|$) and edges ($|\tilde{E}^{\ast}|$ in the network and also an estimate about average number of nodes which can be influenced in the network (irrespective of the start state) ($\tilde{I}^{\ast}$). We now describe how we use these estimates to design our heuristic to compute the goal value for each start state. 
\begin{figure}
  \centering
  \includegraphics[width=0.9\linewidth,height=1.7in]{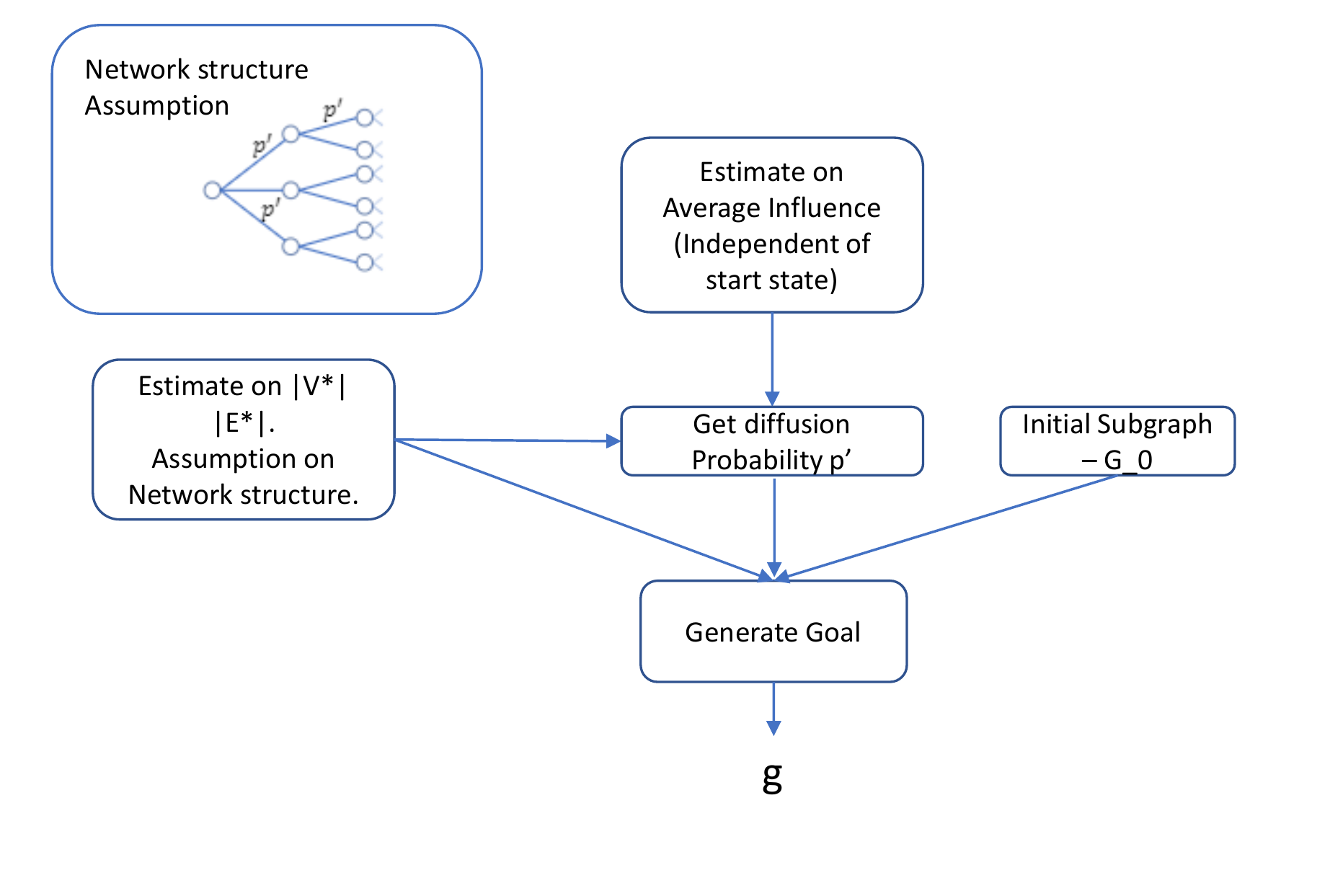}
  \vspace{-0.1in}
  \caption{Process to generate the goal for each start state}
  \vspace{-0.1in}
  \label{fig:goalgen}
\end{figure}

Figure \ref{fig:goalgen} represents the steps for our heuristic. As the network is unknown to the algorithm, we assume a network structure and compute the diffusion probability based on the assumed network structure and estimates about the number of nodes, edges and average influence. By using the computed diffusion probability, given estimates and assumed network structure, we generate a goal value for a given initial subgraph.

We assume that the network is undirected and uniformly distributed, i.e, each node is connected to $\frac{2*|\tilde{E}^\ast|}{|\tilde{V}^\ast|}$ nodes. We also assume a local tree structure as shown in Figure~\ref{fig:goalgen}~\footnote{These simplified assumptions work well to approximate the influence propagation. We also observe in our experiments that our heuristic outputs a value which is closer to actual value.} to approximate the actual expected influence within the social network \citep{chen2010scalable,wang2012scalable}. The root of the tree can be any of the $|S|$ nodes (initially given nodes) and each node will be part of only one of such trees. The influence propagation probability is assumed to be $p'$ and is considered same for all edges. We now show how the value of $p'$ can be computed based on the network structure assumption and available information.
\begin{enumerate}
    \item \textbf{Computing p':} We find a value of $p'$ such that the expected influence in our tree-structured network is similar to the estimate on the average value of influence $\tilde{I}^{\ast}$. To compute the expected influence or expected number of nodes activated in the network, we need to know the number of layers in the tree structure. Therefore, we first compute the number of layers in our assumed tree structure.  Let $K_{1} = |S|*2*\frac{|E^\ast|}{|V^\ast|}$, which is the number of nodes at first layer. For subsequent layers, each node will be connected to $2*\frac{|\tilde{E}^\ast|}{|\tilde{V}^\ast|} -1$ nodes at the layer below it (one edge will be to the node at the above layer). We use $r$ to denote the quantity  $2*\frac{|\tilde{E}^\ast|}{|\tilde{V}^\ast|} -1$ . As the total number of nodes in the graph is $|\tilde{V}^{\ast|}$, sum of the number of nodes at all layers should be equal to $|\tilde{V}|^{\ast}$. Let $L$ denotes the number of layers. Then,   
{\small
\begin{align}
    &|\tilde{V}|^{\ast} = |S| + K_{1} + K_1 * r + K_{1} * r^2 + .. + K_{1} * r^{L-1} \\
    & \implies \frac{(|\tilde{V}|^{\ast} - |S|)}{K_{1}} = \frac{r^L - 1}{r-1} \label{eq:layer}
\end{align}
}
Solving for $L$ gives $L = \log_{r} (1+\frac{ |\tilde{V}|^{\ast} - |S|)*(r-1)}{K_{1}})$. Now, we compute the expected number of nodes activated (influenced) in our assumed network with the propagation probability $p'$. Let $J$ denotes the expected number of nodes influenced in the network. Then, 
{\small
\begin{align}
    &J= |S| + K_{1}*p' + K_1 * r * p'^2    + K_{1} * r^2 *p'^3\\ \nonumber 
    &\hspace{1in}+ .. + K_{1} * r^{L-1} * p'^{L} \\
    & \implies \frac{(J - |S|)}{K_{1}} = \frac{p'*((p'*r)^L - 1)}{p'*r-1} \label{eq:inf}
\end{align}
}
If our assumed network is similar to actual network, the value of $J$ should be close to $\tilde{I}^{\ast}$, i.e.,  the average number of nodes influenced in the network. Therefore, to find the value of $p'$, we perform a search in the probability space and use the value of $p'$ which makes $J$ closest to $\tilde{|I|}^{\ast}$.

\item \textbf{Computing goal value $g$ for a given initial subgraph:}
Now, to compute the goal value for a given initial subgraph, we use the $p'$ value computed above. The subgraph is known, i.e., the neighbors of nodes in set $S$ ($N_{G^{\ast}}(S)$) are known. Therefore, the number of nodes at first layer is equal to $N_{G^{\ast}}(S)$, i.e., $K_1=|N_{G^{\ast}}(S)|$. For the next layer onwards, we assume a similar tree structure as before with each node connected to $2*\frac{|\tilde{E}|}{|\tilde{V}|}-1$ node at the layer below it. Therefore, to compute the goal value, we substitute $K_1$ as $|N_{G^{\ast}}(S)|$ in equations \ref{eq:layer} and \ref{eq:inf} to compute the number of layers and influence value. We use the value of $p'$ computed above and solve for $J$. The $J$ value obtained is the influence value we can achieve for the given subgraph based on the assumptions and available information. We use the value of $J$ as our goal $g$ for the subgraph.
\end{enumerate}

\subsubsection{Modifications to the MDP formulation:} 
The state and action remain the same as before but due to the introduction of goals, the reward function is now parameterized by the goal. Let $R_{t,g}$ denote the reward obtained at timestep $t$ when the goal is $g$. As we only get episode reward, therefore~\footnote{Normalizing the reward using the goal stabilizes the learning.},  
{\small
\begin{equation}
  R_{T-1,g} = \frac{I_{G^\ast}(O(G_T))-g}{g} ~and~   R_{t,g} = 0, \forall t\neq T-1 \label{eq:reward_func}
\end{equation}
}

\subsection{Algorithm}

\begin{figure*}[t]
    \centering
 \includegraphics[width=0.8\textwidth,height=1.8in]{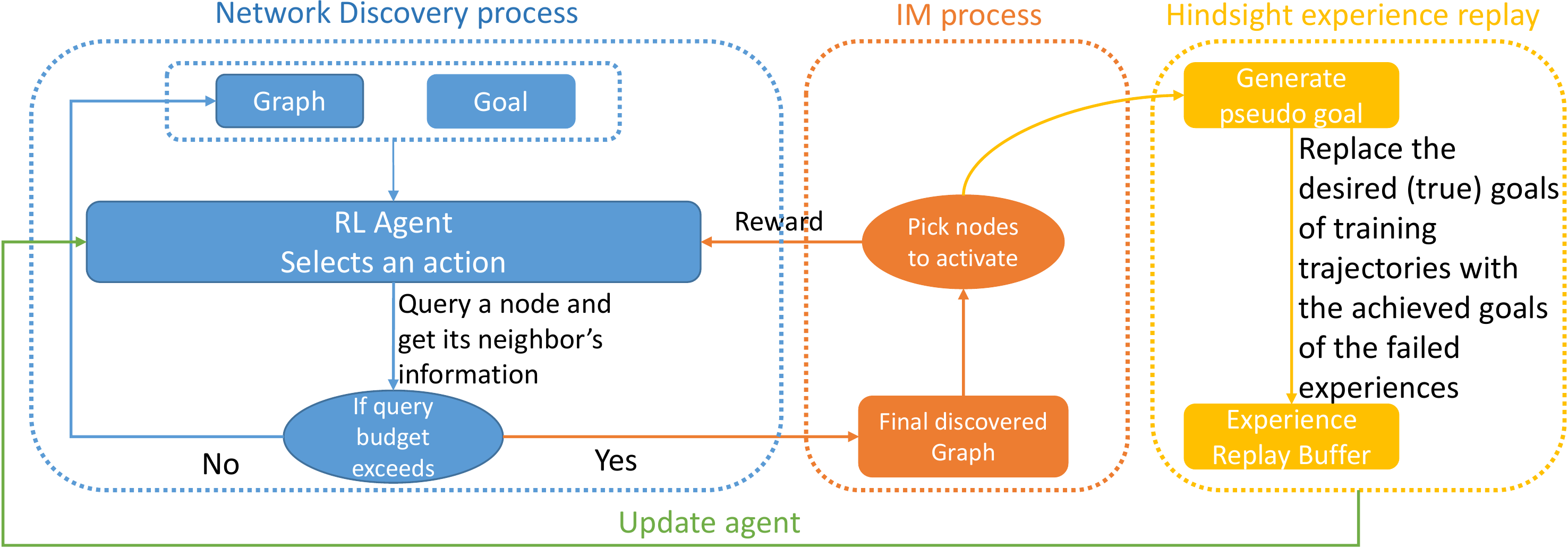}
 \caption{Network discovery framework for influence maximization}
 \label{fig:Structure}
\end{figure*}

In this section, we describe the algorithm used to train the reinforcement learning agent. We use the $DQN$ algorithm and use Curriculum Guided Hindsight Experience Replay for improving the sample efficiency. Algorithm~\ref{al:HER3} describes the detailed steps. Figure~\ref{fig:Structure} provides a visual representation. 

{\small
\begin{algorithm}
\caption{Train Network}
\label{al:HER3}
\SetAlgoLined
\KwIn{Train graphs $\mathcal{G}=\{G_1,G_2,...,G_k \}$, number of episodes $N$. Query budget $T$}
Initialize DQN $Q_{\theta}$ and target DQN $Q_{\theta^\prime}$ with $\theta=\theta^\prime$ and the Replay Buffer $\mathcal{B}$\;
\For{episode $=1$ to $N$}{
            $G=sample(\mathcal{G})$, $S=sample(G)$\;
            Initialize the subgraph $G_0 = (S \cup N_{G^{\ast}}(S) , E(S,N_{G^{\ast}}(S)))$ and corresponding desired goal $g$\;
            $F_0^0=DeepWalk(G_0)$ and $S_0=(F_0^0,A_0)$ \label{lst:line:emb}\;
            Get the possible action set $X= N_G(S)$\;
            \For{$t=0$ to $T-1$}{
                With probability $\epsilon$ select a random node $u_{t}$ from $X$ and with probability 1-$\epsilon$ select $u_t \leftarrow \max_{u \in X} Q_\theta(S_t,\phi(u),g)$ \label{lst:line:query}\;
                Query node $u_t$ and observe new graph $G_{t+1}$\;
                Update the state  $F_{t+1}^0=DeepWalk(G_{t+1})$\ and $S_{t+1}=(F_{t+1}^0 , A_{t+1})$\;
                Update the possible action set $X$  which is the set of nodes not yet queried in $G_{t+1}$\;
            }
            \For{$t=0$ to $T-1$}{
               Store the transition $(S_t,\phi(u_t),R_t,g,  S_{t+1},g)$ in $\mathcal{B}$ \label{lst:expreplay}\;
               Sample the additional goals $\mathbb{G}$ for replay \label{lst:line:sample}\;
                \For{$g^\prime \in \mathbb{G}$}{
                     Add the transition $(S_t,\phi(u_t),R_{t,g^\prime}, S_{t+1},g^\prime)$ to replay buffer $\mathcal{B}$;  \label{lst:line:her}
                }
            }
            \For{$t = 0$ to $T-1$}{
                 Sample a minibatch $A$ from the replay buffer ${\cal B}$(according to the proximity and diversity scores)  \label{lst:line:cher}\;
                  Update the proximity-diversity trade-off parameter $\lambda \leftarrow \gamma \times \lambda$ ;
           
                 Update $Q_{\theta}$ using the minibatch $A$\;
           }
             Update target network $Q_{\theta^\prime}$ with parameters of $Q_{\theta}$ at regular intervals\;
            
}

\end{algorithm}
} 

We train using multiple training graphs. In each episode, we sample a training graph and then sample initial set of nodes $S$. We generate the input state by computing the deepwalk embeddings at each timestep and use $\epsilon-$greedy policy to select the action, i.e., the node to be queried. In step~\ref{lst:expreplay}, we store the transitions according to standard experience replay where we add the goal as well in the experience buffer.

Steps \ref{lst:line:sample}-\ref{lst:line:her} are the \textbf{\textit{first set of crucial steps}} to improve the sample efficiency, where as per the Hindsight Experience Replay technique proposed by \citet{Andrychowicz2017}, we sample pseudo goals and in addition to storing the sample with the actual goal for the episode, we also store each sample by modifying the desired goal (which the agent could not achieve in the failed trajectory) with a pseudo goal $g'$. The reward with the pseudo goal $g'$ is recomputed as per the Equation ~\ref{eq:reward_func}.

While there are multiple possible strategies to generate the set of pseudo goals \citep{Andrychowicz2017}, the most common strategy to generate the pseudo goals is to use the goal achieved at the end of episode. Therefore, in this work, we use $g'$ as $I_{G^\ast}(O(G_T))$. 

\noindent Step \ref{lst:line:cher} is the \textbf{\textit{second crucial step}} towards improving the sample efficiency where for sampling experiences from the replay buffer, we use a curriculum guided selection process which relies on the goal-proximity and diversity based curiosity \citep{Fang2019}. Instead of sampling experiences uniformly, we select a subset of experiences based on the trade-off between  goal-proximity and diversity based curiosity. This plays an important role in guiding the learning process. A large proximity value enforces the training to proceed towards the desired goals, while a large diversity value ensures exploration of different states and regions in the environment. To sample a subset $A$ of size $k$ for replay from the experience buffer ${\cal B}$, the following optimization needs to be solved:
\begin{equation}\label{eq:selection}
\underset{A\subseteq {B}, |A|\leq k}{\max} F(A) =  \underset{A\subseteq { B}, |A|\leq k}{\max} (F_{prox}(A) + \lambda F_{div}(A)) 
\end{equation}

where $B$ is the uniformly sampled subset of size $mk$ from the buffer ${\cal B}$. Let $m=3$  as \citet{Fang2019} does. $F_{prox}(A)$ measures the proximity of the achieved goals $g^\prime$ in $A$ to its desired goal $g$. The second term $F_{div}(A)$ denotes the diversity of states and regions of the environment in $A$. And the weight $\lambda$ is used to balance the trade-off between the proximity and the diversity. The trade-off between the two values is balanced such that  it enforces a human-like learning strategy, where there is more curiosity in exploration in the earlier stages and later the weight is shifted to the goal-proximity.

In our work, we define the proximity as the similarity between goal values and diversity based on the distance between visited states. This is because even though goal values (influence achieved) can be different, the states visited can still be very similar to each other. Formally, to define proximity, we use the difference between achieved goal $g^\prime_i$ and the desired goal $g_i$ as distance and subtract it from a large constant to get the similarity, i.e.,
\begin{align}
F_{prox}(A) &=\sum_{i\in A} (c - |(g_i^\prime-g_i)|) \label{eq:prox}
\end{align}
where $c$ is a large number to guarantee $(c - |(g_i^\prime-g_i)|)\geq 0$ for all possible $g_i$, and $g_i$ is the goal corresponding to experience $i$ in set $A$. 
\noindent For defining diversity, we need to compute similarity between states, and the Geometric DQN architecture allows us to easily compute this value. Diversity is defined as follows 
\begin{equation}
F_{div}(A) = \sum_{j\in {B}} \underset{i \in A}{\max}\{0, sim(s^{emb}_i,s^{emb}_j)\} \label{eq:div} 
\end{equation}
where we use $s^{emb}_i$ to denote the embedding vector of the state (representation of the graph in the embedding space) corresponding to the experience $i$. The embedding vector of the state is the output of the graph convolution and pooling layer (input to FC1) in Figure \ref{fig:gcn-arch}. $sim(s^{emb}_i,s^{emb}_j)$ denotes the similarity score between the vector representations and is computed by taking the dot product of the vectors.

This definition of diversity is inspired by the facility location function  \citep{cornuejols1977uncapacitated,lin2009graph} which was also used by \citet{Fang2019}. Intuitively, this diversity term is measuring how well the selected experiences in set $A$ can represent the experiences from ${B}$. A large diversity score $F_{div}(A)$ indicates that every achieved state in $B$ can find a sufficiently similar state in $A$. A diverse subset  is more informative and thus helps in improving the learning.

It has been shown that $F(A)$ defined in equation ~\ref{eq:selection} is a monotone non-decreasing submodular function~\footnote{It is a weighted sum of a non-negative modular function ($F_{prox}(A)$) and a submodular function ($F_{div}(A)$). Please refer to the paper by \citet{Fang2019} for details.} Therefore, even though exactly solving equation ~\ref{eq:selection} is NP-hard, due to the submodularity property, greedy algorithm can provide a solution with an approximation factor $1-\frac{1}{e}$ \citep{nemhauser1978analysis}. The greedy algorithm picks top $k$ experiences from the buffered experiences $B$. It will start by taking $A$ as an empty set and at each step, it will add the experience $i$ which maximizes the marginal gain. We denote the marginal gain for experience $i$ by $F(i|A)$ and it is given by
\begin{align}
    F(i|A) &= F(i \cup A) - F(A)
\end{align}

Therefore, by using equations \ref{eq:selection}-\ref{eq:div}, we get
\begin{equation}
\resizebox{.99\linewidth}{!}{$
\begin{aligned}
F(i|A) &= (c- (|g_i^\prime-g_i|)) +  \lambda \sum_{j\in {B}}  \max\{0, (sim(s^{emb}_i,s^{emb}_j) \nonumber \\ &- \underset{l \in A}{\max}(sim(s^{emb}_l,s^{emb}_j)))\}
\end{aligned}
$}
\end{equation}

At the end of each episode, the trade-off coefficient $\lambda$ is multiplied by a discount rate $\gamma$, which produces the continuous shifting of weights from diversity to proximity score. Then effect of $F_{div}(A)$ will go to zero when $\lambda\rightarrow 0 $.

\section{Experiments}

The goal of the experiment section is to evaluate the performance of our approach CLAIM in comparison to following state-of-the-art approaches: 
\squishlist
\item \textbf{Random} - At each step, it randomly queries a node from  available unqueried nodes. 
\item \textbf{CHANGE} Algorithm by \citet{Wilder18} 
\item \textbf{Geometric-DQN (Baseline)} Algorithm by \citet{kamarthi2019influence}
\squishend

 \begin{table}[ht]
 \centering
  \begin{tabular}{ccl}
    \hline
        Category & Train networks & Test networks \\
\hline
    Retweet          & copen, occupy  & israel, damascus,\\ 
            & &                            obama, assad                \\
    Animal           & plj, rob       & bhp, kcs\\
    FSW           & zone 1       & zone 2, zone 3\\
  \hline
\end{tabular}
\vspace{-0.1in}
  \caption{Train and test networks}
  \vspace{-0.1in}
  \label{tab:data}
\end{table}

    \begin{table*}[ht]
        \caption{Comparison of influence score of our proposed approach and existing approaches for each test network. For each network, a paired t-test is performed and $\ast$ indicates statistical significance of better performance at $\alpha = 0.05$ level, $\ast\ast$ at $\alpha = 0.01$ level, and $\ast\ast\ast$ at $\alpha = 0.001$ level.}
        \label{tab:Result}
        \vspace{-0.1in}
          \centering
        \begin{tabular}{c|cccc|cc|cc}
        \hline
        Network category      & \multicolumn{4}{c|}{Retweet networks} & \multicolumn{2}{c|}{Animals networks} & \multicolumn{2}{c}{FSW networks}\\     \hline
        Test networks         & israel        & damascus$\ast\ast$       & obama$\ast\ast$          & assad$\ast$         & bhp$\ast\ast\ast$              & kcs$\ast\ast$          &zone 2$\ast$     &zone 3  \\
        OPT value             & 113.9         & 195.8          & 154.7          & 134.2         & 111.9            & 113.4            & 20.98     & 16.40    \\
        Random value &31.17 &84.71 &40.81 &69.44 &36.80 &54.39 & 13.26&12.31 \\
        CHANGE value          & 32.42         & 92.41          & 48.61          & 69.77         & 35.87            & 54.52     & 12.60     & 10.51    \\
        Geometric DQN              & 37.33         & 105.2          & 52.01          & 75.12         & 40.12            & 60.81             & 13.65     & 12.35\\
        CLAIM approach & \textbf{38.55}& \textbf{113.1} & \textbf{54.67} & \textbf{77.49} & \textbf{42.25}   & \textbf{64.58}    & \textbf{13.94}   & \textbf{12.48}\\   \hline
        Improve percent       & 3.27\%        & 7.51\%         & 5.11\%         & 3.15\%       & 5.31\%           & 6.20\%   & 2.12\%           & 1.05\%\\   \hline
        \end{tabular}
    \end{table*}
    \begin{figure*}[ht]
      \centering
      \begin{minipage}[b]{\linewidth}
      \includegraphics[width=0.33\linewidth,height = 1.1in]{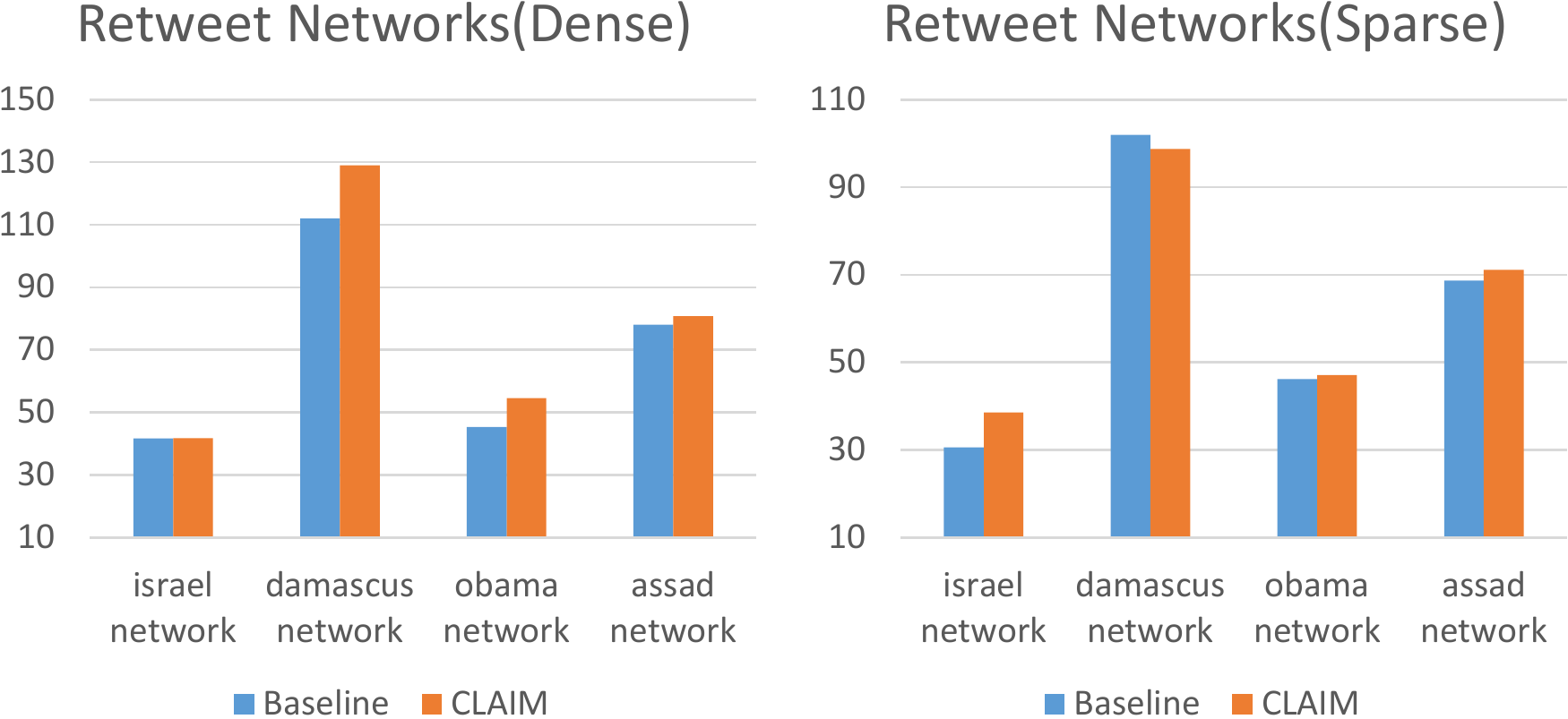} 
      \includegraphics[width=0.33\linewidth,height = 1.1in]{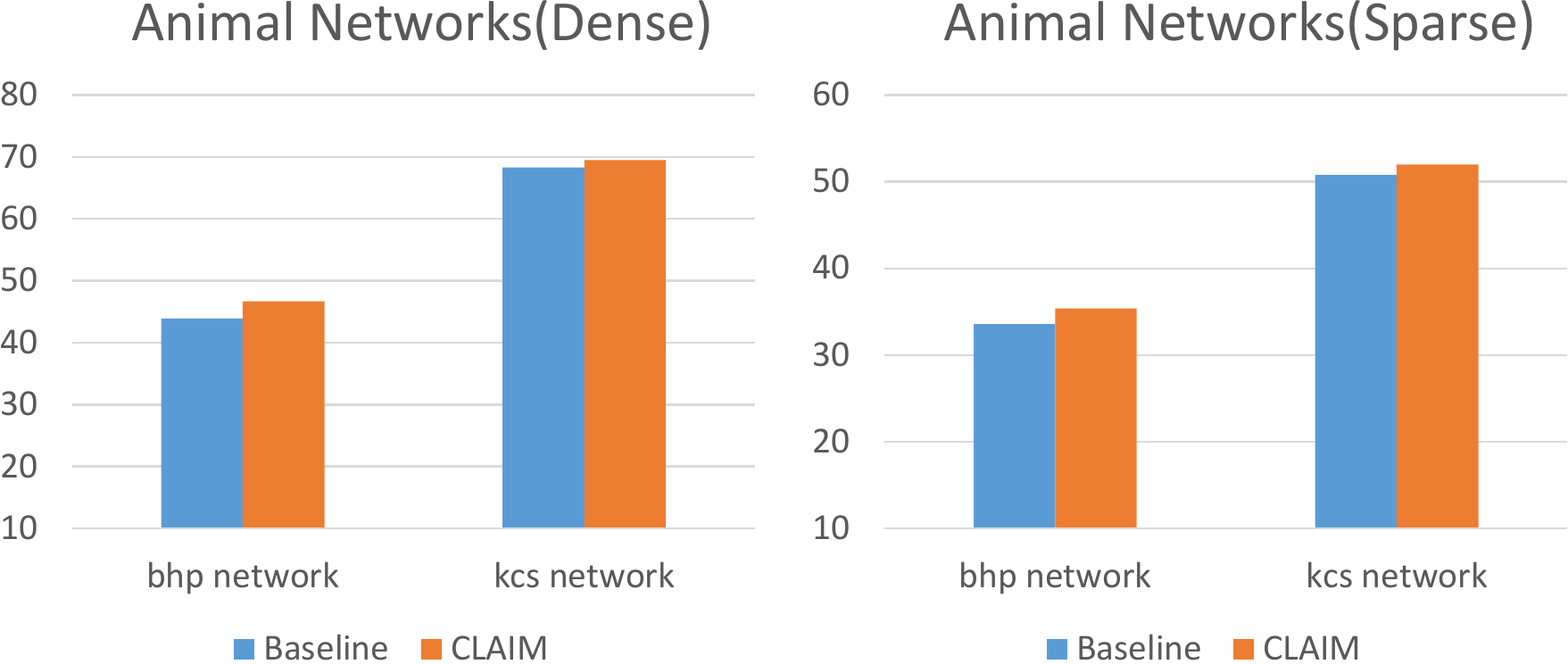} 
      \includegraphics[width=0.33\linewidth,height = 1.1in]{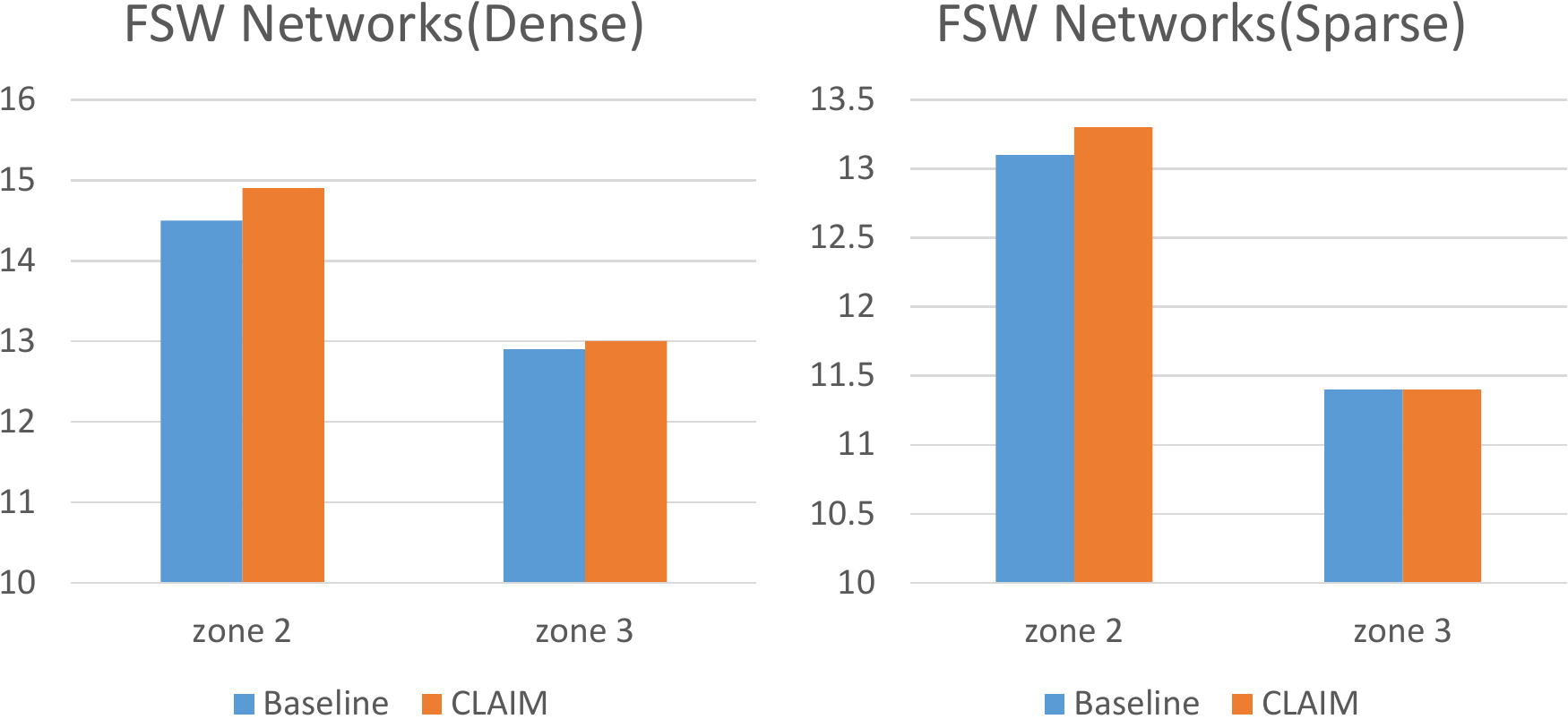}
      \end{minipage}
      \vspace{-0.1in}
      \caption{Comparison of performance of our approach and baseline approach in dense and sparse network environment. }
      \vspace{-0.1in}
      \label{fig:sensity}
    \end{figure*}

\noindent \textbf{Dataset:} The first network is the Retweet Network from twitter \citep{Rossi14}. The second network is Animal Interaction networks which are a set of contact networks of field voles (Microtus agrestis) inferred from mark–recapture data collected over 7 years and from four sites \citep{davis2015spatial}. The third network is a real-world physical network between Female Sex Workers (FSW) in a large Asian city divided into multiple zones. This is a confidential dataset physically collected by a non-profit by surveying different female sex workers recently. The goal in FSW networks is to discover the network and select a subset of FSW from the discovered network to be enrolled in the HIV prevention programs. The enrolled FSWs should be such that they can pass on the information (influence) maximum FSWs in the complete network. For each family of network, we divide them into train and test data as shown in Table \ref{tab:data}.

\noindent \textbf{Experimental Settings:} Our experimental settings are similar to the settings used in \citet{kamarthi2019influence}. There are 5 nodes in the set $S$. All nodes in $S$ and their neighbors are known. We have further budget of $T=5$ queries to discover the network. After getting the final subgraph $G_T$, we pick 10 nodes to activate using greedy influence maximization algorithm. We use $p=0.1$ as the diffusion probability for all the edges.

\subsection{Results}

To demonstrate sample efficiency, we measure the performance of our approach against past approaches by the average number of nodes influenced over 100 runs under a fixed number of queries. Here are the key observations:
\begin{itemize}
    \item \textbf{Average influence value: }Table \ref{tab:Result} shows the comparison of number of nodes influenced by different algorithms. Each algorithm selects the set of nodes to activate from the discovered graph. As shown in the table, our approach consistently outperforms all existing approaches across different networks. CLAIM learns a better policy in the same number of episodes and hence more sample efficient. 
    We would like to highlight here that even a small consistent improvement in these settings is very important as it can ensure more life safety (as an example by educating people about HIV prevention).

    \item \textbf{Effect of density of the initial subgraph:} The number of nodes which can be influenced in the graph is highly dependent on the position of initial subgraph in the whole social network. Therefore, we also test the performance of CLAIM against the baseline approach on the \textit{dense} and \textit{sparse} initial subgraphs (we identify the initial subgraph as dense or sparse based on the ratio of $\frac{|S \cup N_{G^{\ast}}(S)|}{|S|}$). We compare the average influence values as shown in Figure \ref{fig:sensity}. CLAIM outperforms the baseline in most of the cases, except the sparse case in the damascus network. The reason for this may be that the damascus network is an extremely sparse network, and it has some specific structure property that leads this result.

    \item \textbf{Ablation Study:} We also present the detailed results for our ablation study over all datasets in Table \ref{tab:ablation}. We observe the effect of adding each additional component in CLAIM one by one. First, we add only goal as a feature to the baseline model. Next, we add the Hindsight Experience Replay and finally we add the curriculum guided selection of experiences for replay. These results indicate that a single component can not guarantee a better result for all networks, and we need all three components to improve the performance across multiple datasets. 

    \item \textbf{Stability check:} We check the stability of CLAIM by comparing the performance of models trained using different random seeds. We train three models for both baseline and CLAIM. Table \ref{tab:stable} shows the mean and deviation of influence value for different networks. CLAIM not only achieves high mean it also provides a low deviation reflecting the stability of approach.

    \begin{table*}
    \begin{tabular}{c|cccc|cc|cc}
    \hline
    Network category      & \multicolumn{4}{c|}{Retweet Networks} & \multicolumn{2}{c|}{Animals networks} & \multicolumn{2}{c}{FSW  networks}\\     \hline
    Test networks         & israel        & damascus       & obama          & assad         & bhp              & kcs          &zone 2     &zone 3  \\
    \hline
    Baseline (Geometric DQN)          & 37.33         & 105.2          & 52.01          & 75.12         & 40.12            & 60.81             & 13.65     & 12.35\\
    Goal-directed Geometric DQN     & 36.24         & 110.5          & 51.61          & 73.68        & 41.59            & 62.80            & 13.79     & 12.32\\
    Hindsight Experience Replay     & 37.79         & 109.4          & 53.51          & 76.32         & 42.00            & \textbf{64.64}   & 13.81     & \textbf{12.48}    \\
    Our proposed approach (CLAIM) & \textbf{38.55}& \textbf{113.1} & \textbf{54.67} & \textbf{77.49} & \textbf{42.25}   & \textbf{64.58}    & \textbf{13.94}   & \textbf{12.48}\\    \hline
    \end{tabular}
    \caption{Ablation study for each test network}
    \label{tab:ablation}
    \end{table*}
    
            \begin{table}[t]
            \centering
            \begin{tabular}{c|cc}
            \hline
            Networks\textbackslash{}Method & Geometric DQN             & CLAIM         \\ \hline
            israel                           & $37.11\pm 0.42$      & $38.32\pm 0.32$         \\
            damascus                         & $104.2\pm 5.22$      & $112.8\pm  4.01$     \\
            obama                            & $52.15\pm 1.05$      & $54.78 \pm 0.78$     \\
            assad                            & $74.53 \pm 1.71$       & $77.45\pm 1.02$   \\ \hline
            bhp                              & $40.24 \pm 1.25$     & $42.37\pm 0.81$   \\
            kcs                              & $59.67\pm 1.91$     & $63.21 \pm 1.43$    \\ \hline
            zone 2                              & $13.62 \pm 0.01$     & $13.94\pm 0.00$   \\
            zone 3                             & $12.23\pm 0.01$     & $12.45 \pm 0.00$    \\ 
            \hline
            \end{tabular}
            \vspace{-0.1in}
            \caption{Stability of our approach compared to the baseline on different sets of 100 runs}
            \label{tab:stable}
            \vspace{-0.2in}
        \end{table}
    
    \begin{figure}
    \centering
      \begin{minipage}[b]{\linewidth}
            \includegraphics[width = \linewidth,height =1.0in]{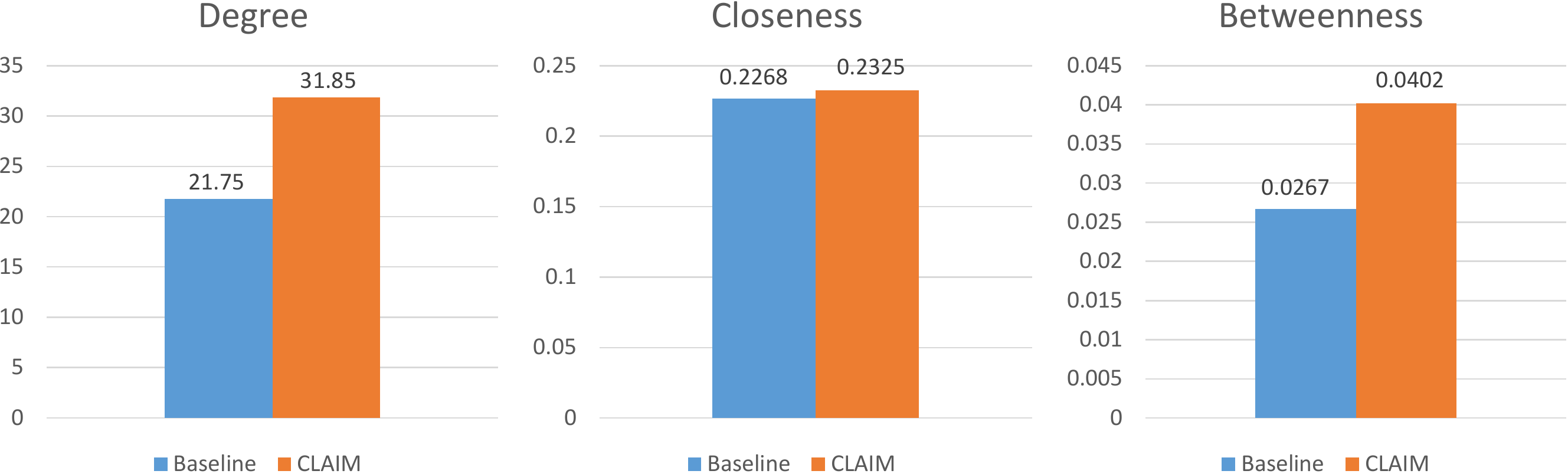}\\
            \includegraphics[width = \linewidth,height=1.0in]{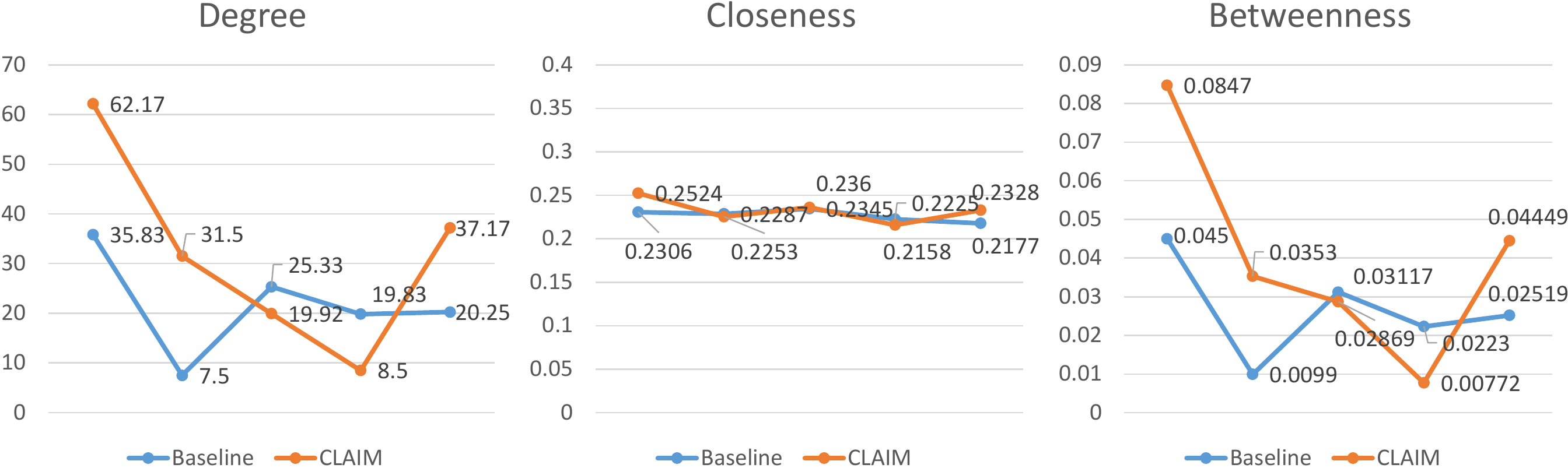}
      \end{minipage}
      \vspace{-0.1in}
        \caption{{\small Top Graph - Average degree, closeness and betweenness centrality of nodes queried in the full graph by CLAIM and baseline. Bottom Graph - Variation of these properties across timesteps.}}
        \label{fig:degree}
        \vspace{-0.15in}
    \end{figure}

    \item \textbf{Property insight:} We also explore the properties of the selected nodes to further investigate why CLAIM performs better. We look at \textit{degree centrality measures, closeness centrality measures,} and \textit{betweenness centrality measures} of the nodes queried in the underlying graph. In particular, we conduct experiment  using \textit{assad}, a retweet network with sparsely interconnected star-graph. As we can see in Figure \ref{fig:degree}, compared to the baseline approach, on an average, CLAIM can recognize nodes with higher degree, closeness and betweenness centrality. As a result, CLAIM is able to discover a bigger network. The higher degree centrality, higher closeness centrality and higher betweenness also show that CLAIM can explore nodes which plays an important role in influence maximization problem. Besides, these values are large at the beginning which means that CLAIM tends to explore a bigger graph first, and then leverage the available information with the learned graph to find complex higher-order patterns in the graphs that enable it to find key nodes during the intermediate timesteps, and finally utilise all the information to expand the discovered graph at the end.

\end{itemize}

\section{Discussion}
We provide a justification for the choices made in the paper.  
\begin{itemize}
    \item \textbf{Network structure assumption for goal generation:} As we have no prior information about the networks except the initial nodes, we need to make some assumption to compute the goal value. We make the assumption of network being uniformly distributed and use a tree structure to approximate the information propagation as most networks  observed for these problems have similar structure or can be converted in these forms with minimal loss of information. 
    \item \textbf{Goodness of heuristic used for goal generation:} Experimentally, we observe that the goal value computed by our heuristic is closer to the actual value. For example, for the training network \textit{copen}, the achieved influence value by the model after training is at most within 20\% of goal value computed using heuristic. In addition, most of the achieved influence value is much closer and is smaller than the computed goal. In the future, we will investigate different ways to generate a goal with proven upper bound. 
\end{itemize}

\section{Conclusion}
In this work, we proposed a sample efficient reinforcemernt learning approach for network discovery and influence maximization problem. Through detailed experiments, we show that our approach outperforms existing approaches on real world datasets. In future, we would like to extend this work to consider multiple queries at each timestep.

\bibliography{li_558}

\appendix
\providecommand{\upGamma}{\Gamma}
\providecommand{\uppi}{\pi}

\end{document}